\documentclass{article}
\usepackage{spconf,amsmath,graphicx}
\usepackage{todonotes}
\usepackage{multirow}
\usepackage{flushend}
\usepackage{url}

\newenvironment{myenum}{
\begin{enumerate}
 \setlength{\itemsep}{1pt}
 \setlength{\parskip}{0pt}
 \setlength{\parsep}{0pt}
}{\end{enumerate}}

\title{\textit{U}-Phylogeny: Undirected Provenance Graph Construction in the Wild}

\name{A. Bharati$^1$, D. Moreira$^1$, A. Pinto$^{1,2}$, J. Brogan$^1$,\textit{ K. Bowyer$^1$, P. Flynn$^1$, W. Scheirer$^1$ and A. Rocha$^{1,2}$}\thanks{This material is based on research sponsored by DARPA and Air Force Research Laboratory (AFRL) under agreement number FA8750-16-2-0173. Hardware support was generously provided by the NVIDIA Corporation. We also thank the financial support of FAPESP (Grant \#2015/19222-9),  CAPES (DeepEyes Grant) and CNPq (Grant \#304472 /2015-8).}}

\address{$^1$Department of Computer Science and Engineering, Univ. of Notre Dame, IN, U.S.A.\\$^2$ Institute of Computing, Univ. of Campinas, SP, Brazil}

\begin{document}
\ninept
\maketitle

\begin{abstract}
Deriving relationships between images and tracing back their history of modifications are at the core of Multimedia Phylogeny solutions, which aim 
to combat misinformation through doctored visual media. Nonetheless, most recent image phylogeny solutions cannot properly address cases of forged composite images with multiple donors, an area known as multiple parenting phylogeny (MPP). This paper presents a preliminary undirected graph construction solution for MPP, without any strict assumptions. The algorithm is underpinned by robust image representative keypoints and different geometric consistency checks among matching regions in both images to provide regions of interest for direct comparison. 
The paper introduces a novel technique to geometrically filter the most promising matches as well as to aid in the shared region localization task. The strength of the approach is corroborated by experiments with real-world cases, with and without image distractors (unrelated cases).
\end{abstract}
\section{Introduction}
\label{sec:intro}
One key concern in digital forensics nowadays is how to fight propaganda and misinformation through visual media. With online visual content easily accessible, their reuse and iterative upload and download naturally lead to the presence of multiple copies of a single object. These copies can be generated through a series of transformations solely from an original image $r$ (so-called near duplicates), from different originals $r_i$ and $r_j$ depicting nearly the same scene but each of which with its own chain of modifications (so-called semantically-similar images) or be combined with various other image donors $d_i$ (generating composite images). Dias et al.~\cite{Dias_2012} studied near-duplicate images and proposed a method to find their kinship relationships or the directions of modifications (and transformations) over time, terming such analysis as image phylogeny. Semantically-similar images were studied in a follow-up work~\cite{Dias_2013}. 

\textcolor{black}{Extending those works, Oliveira et. al~\cite{Oliveira_2014} formalized cases of image forgeries and compositions with what they called Multiple Parenting Phylogeny (MPP). In an MPP setup, an image can be derived from multiple donors and thus its content might have common pieces with all those donors. Moreover, each composite and donor image might have its own chain of near duplicates and semantically-similar images. However, the best MPP solution that exists to date works only with bi-composite images (images that have two parents --- a host and a donor image), which solves only a restricted case of MPP, leaving the more difficult general problem still largely untouched. With donors from multiple images, the information required to estimate the correct transformations to map an image onto each of its possible donors, might not be present, rendering existing MPP solutions inadequate for use in such cases.}

\begin{figure}[t]
\begin{center}
	\includegraphics[width=8cm]{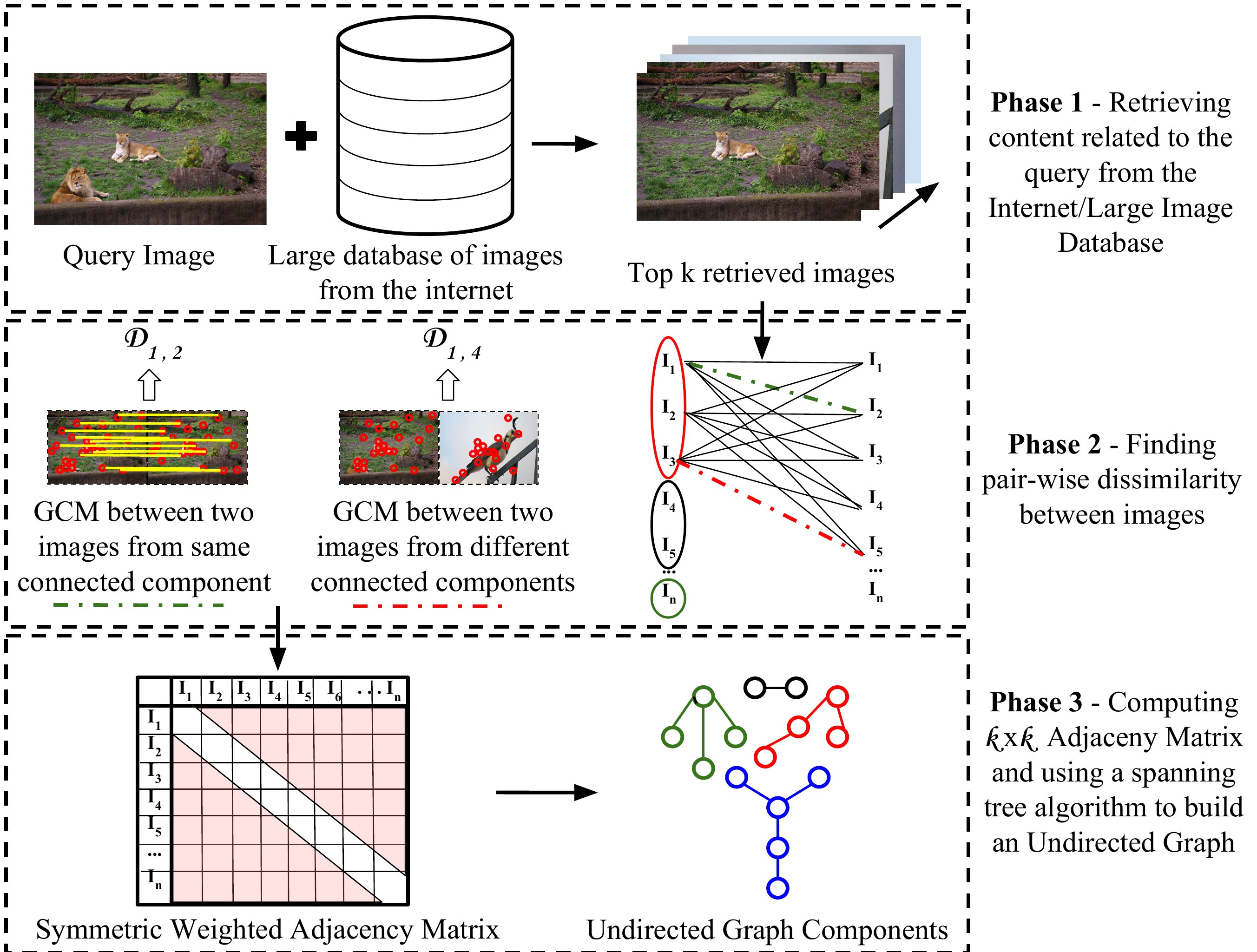}
	\caption{Different stages of the proposed pipeline for provenance graph building of a given query and its possible donors. Upon searching an image collection and retrieving possible donors, we compare candidates pairwise through the matching of representative keypoints, geometrically check the consistency among possible matches and build a weighted dissimilarity matrix representing all possible pairwise relationships. Ultimately, we use a spanning tree to find the connected components related (and unrelated) to the query. \textbf{Note:} GCM stands for geometrically consistent matches. \label{fig:pipeline}}
	\end{center}
    \vspace{-0.9cm}
\end{figure}

\begin{figure}[t]
\begin{center}
	\includegraphics[width=8.0cm]{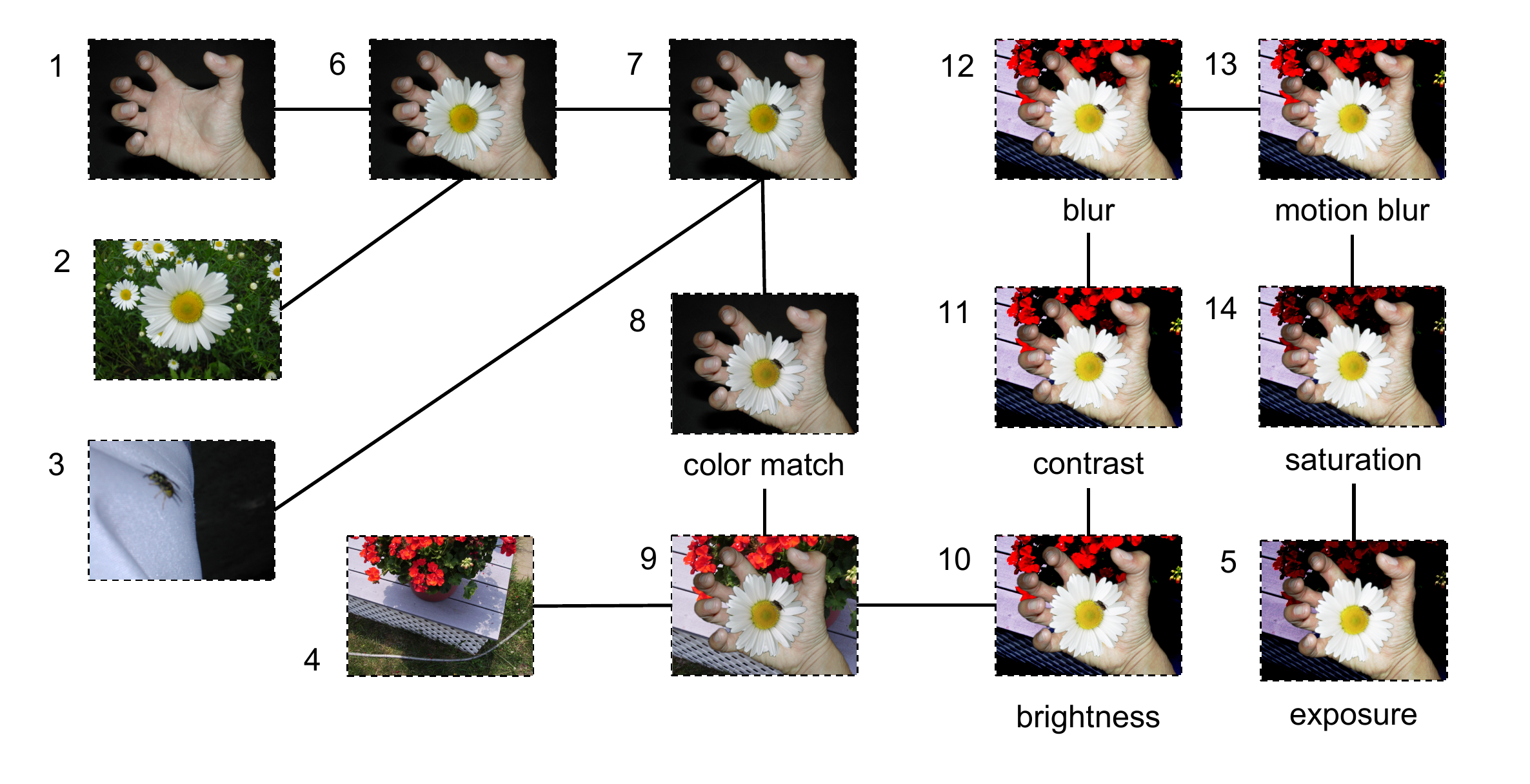}
	\vspace{-0.2cm}
	\caption{Undirected phylogeny graph for the case of bi-composite (e.g., node 6) and multi-composite images (e.g., node 7). Examples from~NIST NC2017 dataset~\cite{Nimble_2017}.\label{fig:nomenclature}}
	\end{center}
	\vspace{-0.8cm}
\end{figure}

Multimedia phylogeny solutions can be useful in detecting visual media frauds, preventing propaganda and misinformation dissemination, and resolving news media controversies. Therefore, it is important to devise methods that generalize to different image transformations as well as to any number of possible donors. Updating some terminology used in~\cite{Oliveira_2014} and \cite{Oliveira_2016}, for this work, rather than differentiating a donor as a possible host (donor of the background of a composite) or alien (additional donors), all images contributing to a composite image are simply referred to as donor images (DIs). The composite image is called a multi-composite image (MCI) (See Fig. ~\ref{fig:nomenclature}).

\textcolor{black}{Inferring directions to connections in the phylogeny graph of an unrelated set of images is difficult as the irrelevant images add noise to the process. Building an undirected graph first helps to reduce the uncertainty in direction finding and is more efficient (in terms of number of image comparisons) for large sets of images than directed phylogeny graph construction. Once the connections are obtained, localized techniques can be devised for pointing out the directions; human experts may also be an option.}

In this paper, we propose an algorithm for Phylogeny Graph Construction that aims at building an undirected graph showing the relationships among images using spatial information provided by representative keypoints and the consistency of their matches. The method eliminates some assumptions made in prior work and generalizes to any set of images, hence "in the wild". The paper also presents results on a difficult dataset recently released by the National Institute of Standards and Technology (NIST) and proposes new metrics for evaluation of Image Phylogeny tasks. \textcolor{black}{Instead of replacing all components of the existing work~\cite{Oliveira_2016}, we build upon them with generalization in mind and extend specific pieces of the method to deal with multiple donors and any kind of image transformation. Fig. \ref{fig:pipeline} shows the end-to-end pipeline we propose. For this paper, we assume an efficient system for Phase 1 (retrieving images from a collection) and use its result (top-$k$ related images to the query) as input for our algorithm.}

\vspace{-5pt}
\section{Related Work}
\label{sec:related}
Finding parental relationships between images was first explored by Kennedy and Chang~\cite{Kennedy_2008} with Visual Migration Maps (VMMs) used to select images of interest from a given set of candidates. The main problem with VMMs, however, is the need of detectors for each image transformation, constraining the representation power of the graph to the detectors in place. In turn, De Rosa et. al~\cite{DeRosa_2010} compared pairs of images using both the image content and the noise information to find possible dependencies. 

The Multimedia Phylogeny term was introduced by Dias et al.~\cite{Dias_2010}  in their work with image phylogeny tree reconstruction with near-duplicate images. Subsequent work introduced solutions for multiple trees~\cite{Dias_2012} and multiple trees with semantically-similar images~\cite{Dias_2013}. None of the works, however, considered images with multiple parents, and so their solutions were in the form of trees. 

Phylogeny (or provenance) graph construction for a more prevalent case of forgery in which objects from one image are spliced into another, was addressed by Oliveira et al.~in their recent papers on multiple parenting phylogeny~\cite{Oliveira_2014,Oliveira_2016}, which are the most relevant to our work herein. The authors propose a solution with a strict assumption of two parents (one host and one alien) of a composite image, and also with no unrelated images in the set. Moreover, the authors assume a fixed set of possible transformations that allow an image to be considered as the near duplicate of another — resampling, cropping and affine transformation, contrast, brightness, gamma correction, and compression (as defined in \cite{Dias_2012}). 

Existing image phylogeny methods mainly focus on two steps to find the provenance graph: (i) computing the dissimilarity matrix for all images in a collection; and (ii) building a directed graph using a spanning tree algorithm. Step (i) is further divided in (a) detecting matching keypoints for every pair of images; (b) estimating the best geometric transformation between those sets of points and warping one image onto the other; (c)  matching color and compression parameters between the pairs of images; and (d) computing a pixel-wise difference between the mapped images. 

While Step~(ii) has been ``the'' subject of research in prior work,  Step~(i) has been overlooked, which has streamlined the research in the field by far. The many constraints with the existing solutions obfuscate the difficulties of the general MPP problem. Firstly, the set of transformations is not exhaustive and there can be transformations in real-world cases that have not been accounted for in the existing literature. \textcolor{black}{Secondly, the methods used for estimating the transformations might have some limitations since they are based on local pixel information. Mapping the color distribution of content-related regions, for instance, can be similar in both directions (e.g., with reverse-prone transformations), thus not proving helpful or discriminatory for kinship direction finding.} 
In addition, the information required to perform compression mappings, such as the compression table, might not be available for non-JPEG lossless-compressed images. Finally, there can be images completely unrelated (in terms of sharing one scene but with similar color distribution) to the query, as part of the result of the retrieval or the effect of the semantic gap~\cite{Datta_2008}. 





\vspace{-5pt}
\section{Proposed Solution}
\label{sec:algo}

\begin{table*}[t]
\caption{Experiments without distractors using \textit{U}-Phylogeny (proposed algorithm) and its different forms of calculating the dissimilarity matrix. $MSE$ denotes Mean Squared Error and $GCM$,  Geometrically Consistent Matches.}
\hspace{2em}
\centering
\begin{scriptsize}
\begin{tabular}{ |c|c|c|c|c|c|c|c|c|c|  }
 \hline
 \multicolumn{10}{|c|}{\textbf{Performance Without Distractor Images}} \\
 \hline
 \multirow{2}{9em}{Dissimilarity Metric} & \multicolumn{3}{|c|}{$Precision_{edges}$}  &\multicolumn{3}{|c|}{$Recall_{edges}$}&\multicolumn{3}{|c|}{$VEO$}\\ \cline{2-10}
   & Small & Medium & Large & Small & Medium & Large & Small & Medium & Large\\
 \hline
 
 Avg. Distance of $GCM$    & 0.62 $\mp$0.20 & 0.47 $\mp$0.08 & 0.31 $\mp$0.16 & 0.62 $\mp$0.20 & 0.48 $\mp$0.08 & 0.32 $\mp$0.16 & 0.82 $\mp$0.09 & 0.75 $\mp$0.04 & 0.66$\mp$0.08\\
 Number of $GCM$ & 0.75 $\mp$0.19 & 0.61 $\mp$0.13 & 0.52 $\mp$0.15 & 0.75 $\mp$0.19 & 0.61 $\mp$0.12 & 0.54 $\mp$0.15 & 0.88 $\mp$0.09 & 0.81 $\mp$0.06 & 0.77 $\mp$0.07\\
 
 $MS$E& 0.73 $\mp$0.19 & 0.56 $\mp$0.10 & 0.42 $\mp$0.04 & 0.73 $\mp$0.19 & 0.56 $\mp$0.10 & 0.43 $\mp$0.03 & 0.87 $\mp$0.09 & 0.79 $\mp$0.05 & 0.72 $\mp$0.02\\
  Mutual Information& 0.76 $\mp$0.17 & 0.64 $\mp$0.16 & 0.57 $\mp$0.12 & 0.76 $\mp$0.17 & 0.65 $\mp$0.16 & 0.58 $\mp$0.11 & 0.89 $\mp$0.08 & 0.83 $\mp$0.08 & 0.79 $\mp$0.06\\

 \hline
\end{tabular}
\end{scriptsize}
\label{tab:res1}
\vspace{-0.2cm}
\end{table*}

\begin{table*}[t]
\caption{Experiments with distractors. $MSE$ stands for Mean Squared Error and $GCM$ stands for Geometrically Consistent Matches.}
\hspace{4em}
\centering
\begin{scriptsize}
\begin{tabular}{ |c|c|c|c|c|c| }
 \hline
 \multicolumn{6}{|c|}{\textbf{Performance With Distractor Images}} \\
 \hline
 Dissimilarity Metric & $Precision_{nodes}$ & $Recall_{nodes}$ & $Precision_{edges}$ & $Recall_{edges}$ & $VEO$\\ 
 \hline
 
 Avg. Distance of $GCM$ & 0.98 $\mp$0.05 & 1.00 $\mp$0.00 & 0.56 $\mp$0.16 & 0.55 $\mp$0.18 & 0.79 $\mp$0.07\\
 Number of $GCM$ & 0.98 $\mp$0.05 & 1.00 $\mp$0.00 & 0.72 $\mp$0.15 & 0.69 $\mp$0.16 & 0.85 $\mp$0.07 \\ 
 $MSE$ & 1.00 $\mp$0.00 & 1.00 $\mp$0.00 & 0.69 $\mp$0.14 & 0.64 $\mp$0.11 & 0.84 $\mp$0.06 \\
 Mutual Information & 1.00 $\mp$0.00 & 1.00 $\mp$0.00 & 0.78 $\mp$0.15 & 0.72 $\mp$0.12 & 0.88 $\mp$0.06 \\

 \hline
\end{tabular}
\label{tab:res2}
\end{scriptsize}
\vspace{-0.2cm}
\end{table*}

The proposed solution points out the undirected binary relations that might exist among the elements of a given set of images, based on their visual content.
These relations aim at supporting the revelation of the phylogeny of the images.
As explained in Sec.~\ref{sec:related}, prior work made strong assumptions regarding the probable phylogeny of the images, advancing the state of the art up to the particular case of donor-host composites.
We extend the literature~\cite{Oliveira_2016} toward the direction of analyzing more general composite cases, here called MCIs.

Fig.~\ref{fig:pipeline} outlines the main steps of the proposed solution.
An end-to-end implementation starts with a query image $q$ of interest, whose donors (if any) are to be discovered, together with possible near-duplicates of the query and the donors.
The first step involves querying a large collection of images, for finding and sorting a list of the top-$k$ potentially related items, according to their similarity to the query (c.f., \emph{Phase 1} in Fig.~\ref{fig:pipeline}, and Sec.~\ref{ssec:retrieval} for details).
Once the top-$k$ related images to the query are retrieved, in \emph{Phase 2}, we calculate the dissimilarity of each pair of images, including the query. For this step, we introduce novel strategies for computing dissimilarities and constitute \emph{symmetric weighted adjacency matrices}, which are robust to varied image transformations, and rely upon image-pairwise keypoint detection, description, and geometrically consistent matching (GCM). This method is termed as \textit{U}-Phylogeny (Undirected Phylogeny). We also propose an extended \textit{U}-Phylogeny by computing dissimilarity values using pixel-wise local dissimilarity computations after the GCM, as this might be useful in some situations, but at the cost of an increased runtime (Sec.~\ref{ssec:matrix}). This method can improve the results since it has more knowledge about forgery through transformations. Ultimately, in \emph{Phase 3}, we estimate the query's provenance graph as a minimally connected undirected subgraph, which can be presented to an expert, or be fed to a further forensic image provenance oracle tool (Sec.~\ref{ssec:phycon}).

%

\vspace{-5pt}
\subsection{Image Retrieval}
\label{ssec:retrieval} 
Retrieving images related to the query image is the first step in the end-to-end pipeline of generating a phylogenetic graph. Different context-based information retrieval algorithms~\cite{Ke_2004, Dong:ICMR:2012, Yuan:ICIP:2015} can be adapted to tackle this part of the problem given large collections. In this paper, we do not focus on this particular task and assume to have been provided with a set of images after retrieval. The top-retrieved images may or may not be related to the query. Prior work on multiple parenting phylogeny~\cite{Oliveira_2014, Oliveira_2016} did not consider unrelated images thoroughly. 

\vspace{-5pt}
\subsection{Computation of the Dissimilarity Matrix}
\label{ssec:matrix}
Given a set of $k$ images, a dissimilarity matrix $D$ is a $k \times k$ matrix with the value of dissimilarity between every pair of images from the set. The matrix can be considered as a weighted adjacency matrix of a graph, in which each image is a node and the values in the matrix correspond to weights of edges between any two nodes. Each edge weight is computed using the following steps:

\vspace{0.1cm}\noindent
\textbf{Detection and Description of Points of Interest.} Speeded Up Robust Features (SURF) \cite{Bay:CVIU:2008} keypoints are detected on both images.
The SURF keypoints highlight the important regions 
within the image content, and provide a description process and representation that are robust to transformations~\cite{tuytelaars_2008}.

\vspace{0.1cm}\noindent
\textbf{Keypoint Matching.} To find correspondences between the detected keypoints in the two images, we compute the matches $M$ between the two sets of descriptors $D_1$ and $D_2$ that are obtained from the previously detected keypoints.
The first set is treated as the query set and the other is treated as the gallery set.
For each descriptor  $d \in D_1$, the best matching descriptor is found inside $D_2$ using L2 distance. In addition, inspired by the Nearest Neighbor Distance Ratio (NNDR) matching quality~\cite{lowe2004distinctive}, we ignore all the keypoints whose ratio of distances to their first and second matched descriptors is smaller than an NNDR threshold $t$, implying that the keypoint might be of poor distinctive quality.


\vspace{0.1cm}\noindent
\textbf{Keypoint Match Filtering.} Once the matches are established, upon using NNDR, it is not uncommon to gather \emph{geometrically inconsistent matches}, thus there is the need to remove spurious matches that are not truly representing the real transformations of one image onto the other (e.g., crossing matches among two images). A contribution of this paper is solving this problem with a filter of matched keypoints that keeps only the matches whose spatial dispositions are geometrically consistent in both images, say $M_{g}$. To obtain $M_{g}$, rather than relying on the value of the matched image pixels, we rely upon the spatial positions of the two best matched points $p_1(x_{p1}, y_{p1})$ -- $p_2(x_{p2}, y_{p2})$ in $I_1$, and $q_1(x_{q1}, y_{q1})$ -- $q_2(x_{q2}, y_{q2})$ in $I_2$.
By taking the positions, the distance $l_1$, and the angle $a_1$ between $p_1$ and $q_1$ (all from image $I_1$), as well as the positions, the distance $l_2$, and the angle $a_2$ between $p_2$ and $q_2$ (all from image $I_2$), we estimate the constraints with respect to scale, translation, and rotation transformations, from $I_1$ onto $I_2$, thus applying them on all the keypoints of $I_1$.
We then compare the new positions of $I_1$ and the positions of $I_2$, and remove the keypoints (and respective matchings) that do not follow the estimated constraints. 


%

\vspace{0.1cm}\noindent
\textbf{Dissimilarity.} Finally, we compute the dissimilarity \begin{math} D_{1,2} \end{math} between the two images.
For \textit{U}-Phylogeny, we use the number of filtered (geometrically consistent) matches, $M_{g}$ and the average match score 
of these matches as dissimilarity. Upon filtering the keypoint matches, 
for the extended (more expensive) algorithm, a few more steps are involved before computing the dissimilarity. The keypoints corresponding to the filtered matches are used to estimate homography between the two images. The images are registered based on the estimated parameters. Localized regions of interest (ROIs) are cropped from the registered images by computing the bounding box of the convex hull around the filtered keypoints. 


The pixel value distribution of the two ROIs is matched using a frequency-based histogram-matching approach. The method involves computing the cumulative distribution for the pixel values for both source and target images (in each image pair). Each pixel of the source image is mapped onto the closest pixel value from the same quantile of the target histogram. Since our dissimilarity matrix is symmetric, the transformation only needs to be performed once (either of the two images can be a source image). Then, we compute the pixel-wise dissimilarity, in the form of Mutual Information and MSE, between the mapped source image and the target image. 
\vspace{-5pt}
\subsection{Phylogeny Graph Construction}
\label{ssec:phycon}
Upon obtaining the complete dissimilarity matrix, Kruskal's Minimum Spanning Tree (MST) algorithm \cite{Kruskal:AMS:1956} is used to build an undirected graph connecting all images.
The method requires two inputs -- the number of retrieved images and the weighted adjacency matrix containing real-valued finite weight for each edge. The output graph is a binary adjacency matrix ($BAM$) for which $BAM_{ij}$ is set to 1 whenever there is an edge (i.e., $edge(i, j) \in MST$).

\section{Experimental Setup}
\label{sec:exp}

\noindent
\textbf{Datasets Used.} For evaluation, we use NC2017-DEV2 dataset provided by the National Institute of Standards and Technology (NIST) as part of the Nimble 2017 Challenge~\cite{Nimble_2017}. The dataset is divided into query set (59 images) and gallery set (10446 images). Images from the gallery set may or may not be related to the query set. The dataset has 59 phylogeny cases with 750 images in total. The average graph order (i.e., the average number of related images) for such cases is 12.7. The range of number of related images is \begin{math} [3, 82] \end{math}. 
We organized such cases into three categories based on the number of nodes --- small ($\leq$12 nodes), medium (13-20 nodes), and large graphs ($>$~20 nodes). 

To evaluate the robustness of the methods, we consider building the graphs under the presence of unrelated images. For this particular experiment, we sample 20 cases with graph order $\leq 25$. We process 25 images each time, regardless of the variable size of the provenance graphs. For instance, if we have a test case with 5 nodes, we complete this case with 20 randomly selected distractors and perform the analysis considering 25 nodes in total. The materials for reproducing this work are available at \url{https://gitlab.com/notredame-provenance/u-phylogeny}. 

\vspace{0.2cm}\noindent
\textbf{Evaluation Metrics.} Existing metrics for evaluating image phylogeny  generally focus on the notion of image phylogeny trees~\cite{Dias_2012} and do not conform with undirected graphs as there is no notion of roots, leaves or ancestors therein. Hence, we rely on more general graph comparison metrics to evaluate results. We use precision and recall of nodes and edges and a combined metric, Vertex and Edge Overlap (VEO) as discussed in \cite{Papadimitriou_2010} in the context of web graphs. For each provenance case, the values for these metrics are obtained by comparing the output graph $G'$ of a method with the ground truth graph $G$ using the formulae in Eqs.~\ref{eqn:prec} and \ref{eqn:veo}, where $P$ and $R$ stand for precision and recall, respectively. Here, $nd_{G'}$ denotes the set of nodes (images) in graph $G'$ while $nd_{G}$ denotes the same for graph $G$. Precision and recall of edges is computed similarly.
\begin{equation}
\begin{scriptsize}
\label{eqn:prec}
    P(nd) = \frac{|nd_{G'} \cap nd_{G}|}{|nd_{G'}|}
    \hspace{0.4cm}
    R(nd) = \frac{|nd_{G'} \cap nd_{G}|}{|nd_{G}|}
\end{scriptsize}
\vspace{-0.2cm}
\end{equation}

\begin{equation}
\begin{scriptsize}
\label{eqn:veo}
\begin{aligned}
    VEO(G', G) & = & 
    2*\frac{|nd_{G'}\cap nd_{G}|+|edges_{G'}\cap edges_{G}|}{|nd_{G'}|+|nd_{G}| + |edges_{G'}|+|edges_{G}|}
\end{aligned}
\end{scriptsize}
\end{equation}

The metrics take values in the range of $0$ to $1$. Higher values indicate better performance. The overall values reported for these metrics have been averaged over the three categories in the dataset. 

\vspace{0.2cm}\noindent
\textbf{Experimental Details.} Upon receiving a list of images for  phylogeny graph construction, we can have both related and unrelated images and we need to refine the list to create the graph. With this in mind, we divided the experiments into two setups: 

\begin{myenum}
\item \textbf{Without Distractor Images.} In this setup, all analyzed images for phylogeny graph construction are related to the query image. Differently from prior work~\cite{Oliveira_2016}, there might be multiple donors for each given query.  

\item \textbf{With Distractor Images.} This scenario comprises possible failures in the retrieval of possibly related images and evaluates the performance of \textit{U}-Phylogeny and its extensions in the presence of related and unrelated images.
\end{myenum}

For the dissimilarity matrix, we compute 2000 SURF keypoints for each image. The quality of these keypoints is governed by the hessian threshold (set to 100 to select the most important keypoints) and NNDR (individually computed for each image based on the top two-matched keypoints). 
Following~\cite{lowe2004distinctive}, we use an NNDR threshold $t = 0.8$.
The detected keypoints are filtered using the three parameters of rotation, scale and translation, individually computed for each image. For the $U$-phylogeny version, we use the match distance and count of these keypoints as dissimilarity.

For the extended version, the images are registered using the affine transformation matrix estimated by these keypoints. After cropping the regions of interest (ROI) and mapping the pixel distribution of one to another, the mutual information and MSE is computed. These become the values of dissimilarity between the images.



\section{Results}
\label{sec:results}

Table~\ref{tab:res1} shows the result of Experiment~1. The $Precision_{nodes}$ and $Recall_{nodes}$ values are not valid for this setup as it has no distractors. Observe that the version of our approach using the keypoint information is on par with the extended version for small phylogeny cases and slightly below par for medium and large cases. It is important to note that the extended version of our algorithm matches the pixel color distribution as an additional step for each pair of images. The results show that the transformation mapping and pixel-wise comparison improve upon $U$-Phylogeny. In addition, mutual information of mapped pixel values outperforms the mean-squared-difference. This result is consistent with the literature~\cite{Oliveira_2016}. The best performance for cases without any distractors is obtained with mutual information as the dissimilarity metric for the extended $U$-Phylogeny. Directly comparing our methods with~\cite{Oliveira_2016} is not possible as~\cite{Oliveira_2016} assumes a fixed set of transforms and at most two donors.

The `in-the-wild' evaluation of the methods uses the same phylogeny cases  with added unrelated images (distractors). Table~\ref{tab:res2} shows the results for this experiment. In this case, we also report the $Precision_{nodes}$ and $Recall_{nodes}$, which represent the methods' performance for connecting images that are related and leaving out the unrelated ones. As can be seen from the values, the proposed approach works remarkably well in getting the connected nodes. $U$-Phylogeny also seems to be robust to distractors in terms of edges and the extended version achieves an 88\% of vertex and edge overlap, a significant result for image phylogeny graph construction. As for efficiency, on an Intel(R) i7-5930K CPU \@3.50GHz with 64GB of RAM, $U$-phylogeny takes 3.8s to compare two images whilst the extended version is twice as expensive.


\section{Discussion on Direction in Phylogeny Graphs and Conclusions}
\label{ssec:direction}
Until now, image phylogeny has been approached as tree-building given an asymmetric dissimilarity matrix. The asymmetry in the values while matching $I_i$ to $I_j$ and $I_j$ to $I_i$ is based on the transformations performed on the source image to match the target image. Assuming the transformations are not symmetric, the pixel-wise dissimilarity values of the two images are different. However, under real-world conditions, more complex (and not necessarily asymmetric) transformations 
might be present, especially when considering multiple donors to a composite. For such cases, the established concept of finding directions fails. 

In this vein, this paper provides a generalized extension to the existing solution for multimedia phylogeny~\cite{Dias_2012} and more specifically to the MPP problem~\cite{Oliveira_2016}. We introduce methods to construct undirected phylogeny graphs for multi-composite cases with unknown number of donors. The method generalizes well over images in non-JPEG formats (another constraint present in prior work~\cite{Oliveira_2016}) by not utilizing the loss information from image compression. Moreover, the paper also introduced the usage of new metrics of evaluation for phylogeny, which are adequate for undirected graphs. The proposed methods are reasonably robust to the presence of distractors landing themselves as promising preliminary solutions for phylogeny graph construction in the wild. Future work will be devoted to further refine the edge connections and infer the kinship directionality. 

\bibliographystyle{icip-bib}
\bibliography{references}

\end{document}